\documentclass{article}

\PassOptionsToPackage{numbers, compress}{natbib}
\usepackage[preprint]{neurips_2026}
\usepackage{neurips_2026}

\usepackage[table]{xcolor}

\usepackage[utf8]{inputenc} % allow utf-8 input
\usepackage[T1]{fontenc}    % use 8-bit T1 fonts
\usepackage{hyperref}       % hyperlinks
\usepackage{url}            % simple URL typesetting
\usepackage{booktabs}       % professional-quality tables
\usepackage{amsfonts}       % blackboard math symbols
\usepackage{nicefrac}       % compact symbols for 1/2, etc.
\usepackage{microtype}      % microtypography
\usepackage{xcolor}         % colors

\usepackage{booktabs}
\usepackage{caption}
\usepackage{multirow}
\usepackage{amsmath}
\usepackage[pdftex]{graphicx} 
\usepackage{titletoc}
\usepackage[page,header]{appendix}
\usepackage[normalem]{ulem}
\title{Fast Training of Mixture-of-Experts for Time Series Forecasting via Expert Loss Integration}

\author{
Btissame El Mahtout $^{1,2}$ \quad
Florian Ziel$^{1}$ \\
$^1$ Chair of Data Science in Energy and Environment, University of Duisburg-Essen, Germany \\
$^2$ Chair of Econometrics and Statistics, Technical University of Dortmund, Germany\\
\texttt{btissame.el@tu-dortmund.de}  \quad \texttt{florian.ziel@uni-due.de}
}

\begin{document}

\maketitle

\begin{abstract}
We propose a novel Mixture-of-Experts framework, referred to as Expert Loss Integration MoE (EliMoE), for time series forecasting. EliMoE addresses the optimization problem arising from small gating weights by incorporating expert-specific losses, which provide each expert with a direct learning signal independent of the gate-assigned weight. Specifically, the overall objective comprises the base forecasting loss and expert-specific losses, allowing individual expert prediction errors to directly influence parameter updates alongside the aggregate forecasting error. The framework also encourages different experts to learn from different temporal segments of the data. The proposed framework is further combined with a partial online learning strategy that enables efficient incremental updates of model parameters. By integrating expert-level loss information with partial online learning, EliMoE improves forecasting performance while retaining computational efficiency. Empirical results across five benchmark datasets from diverse application domains  with different sampling frequencies show that the proposed approach generally outperforms state-of-the-art supervised neural forecasting models, including Transformer-based architectures such as PatchTST, as well as zero-shot time-series foundation models such as TimeMoE. Furthermore, ablation studies confirm the effectiveness of the expert-specific loss integration strategy, highlighting its contribution to enhancing predictive performance.
\end{abstract}

\section{Introduction}
Time series forecasting is a widely used technique for predicting future outcomes, with broad applications across economics, finance, energy systems,  and many other domains. Accurate forecasts are crucial for ensuring efficient capacity scheduling, supporting strategic decision-making, and enabling effective risk management. Even small improvements in predictive accuracy can yield substantial economic value \citep{hyndman2021forecasting}. However, constructing a precise forecasting model is not easily attainable in a world full of uncertainties, as time series data often exhibit seasonality, trends, structural breaks, and other complex patterns. Another important challenge in time series forecasting, besides forecast accuracy, is the computational efficiency of the model, since decisions often need to be made within strict time constraints.

Classical statistical methods such as ARIMA \citep{box_1990_arima} and exponential smoothing \citep{holt2004forecasting} are good benchmarks, particularly for short, well-behaved series. These methods are computationally efficient and often provide reliable forecasting performance. However, they typically rely on restrictive linearity or stationarity assumptions and tend to perform poorly when those assumptions are violated. Modern neural architectures such as recurrent neural networks (RNNs), convolutional models like WaveNet \citep{borovykh2017conditional}, and attention-based Transformers \citep{vaswani2017attention, zhou2021informer} have significantly expanded model capacity and achieved state-of-the-art performance on several forecasting benchmarks. However, these models typically require substantial computational resources and long training times for both model fitting and forecast generation.

Mixture-of-Experts (MoE) architectures could provide a potential solution to the challenges of forecast accuracy and computational efficiency by training multiple expert networks alongside a learned gating mechanism that routes inputs to the most appropriate experts. In practice, however, expert specialization is not guaranteed. When training is driven solely by a global forecasting loss, the gradient signal received by each expert is filtered through the gating weights, which may limit the emergence of distinct expert behaviors. %This issue is particularly pronounced in time-series forecasting, where regime shifts and concept drift require experts to adopt genuinely specialized roles.

We introduce a new MoE framework that jointly addresses the challenges of forecasting accuracy and computational efficiency. We refer to this approach as Expert Loss Integration Mixture-of-Experts (EliMoE). We augment the conventional global forecasting loss with expert-specific loss terms. %, so that each expert receives a direct, undiluted gradient signal tied to its own predictions.
This loss formulation explicitly addresses the weak-gradient problem caused by very small gating weights while encouraging experts to learn from different portions of the time series. We further integrate this objective with a partial online learning strategy, in which the gating network and a subset of expert parameters are updated incrementally as new observations arrive, while shared representational components are refreshed less frequently. This design substantially reduces computational cost while maintaining high forecasting accuracy.

The contributions of this work are as follows. First, we formalize a training objective that combines a global forecasting loss with expert-specific losses, and demonstrate that the proposed objective improves overall forecasting accuracy. Additionally, we apply masking to selected experts to encourage each network to focus on distinct patterns within the data. Second, we combine MoE with a partial online learning scheme that reduces computational cost without sacrificing predictive performance. Third, we evaluate the proposed framework across diverse datasets at multiple sampling frequencies and compare it with competitive neural baselines. Finally, ablation studies isolate the contribution of the expert-specific loss component and confirm its effectiveness in improving forecasting accuracy.

The remainder of the paper is organized as follows. Section 2 reviews related work on time series forecasting and mixture-of-experts models. Section 3 describes the proposed methodology. Section 4 presents the experimental setup and results, and Section 5 concludes.

\section{Related work}
The Mixture of Experts paradigm was introduced by \cite{jacobs1991moe}  as a framework in which specialized experts model different regions of the input space, while a gating network determines which expert to select. \cite{jordan1994hme} later extended this into hierarchical MoE and proposed an expectation–maximization (EM) training algorithm. 

Interest in MoE was revived by \cite{shazeer2017outrageously}, who introduced the sparsely-gated MoE layer. By activating only the top-k experts for each input, they showed that model capacity could scale dramatically without proportional computational cost. They also identified two major routing issues: expert collapse, where only a few experts dominate, and load imbalance, where workloads are unevenly distributed. To address this, they proposed auxiliary load-balancing and importance regularization losses.

Later work extended MoE to large-scale Transformer systems. GShard \citep{lepikhin2021gshard} enabled distributed expert sharding across accelerators, while Switch Transformer \citep{fedus2022switch} simplified routing to top-1 expert selection, reducing communication overhead while maintaining strong performance. 

A recurring challenge in MoE research is maintaining robust expert specialization. Under task loss alone, rarely selected experts receive weak gradients, while heavily used experts risk becoming overly general. Prior work has therefore focused mainly on routing behavior. Load-balancing losses \citep{shazeer2017outrageously, fedus2022switch}, stochastic routing methods such as THOR \citep{zuo2022thor}, BASE Layers \citep{lewis2021base}, and expert-choice routing \citep{zhou2022expertchoice} all improve routing stability and balanced utilization. However, these approaches remain largely routing-focused: they regulate where tokens are routed, but do not directly optimize the predictive quality or specialization of individual experts.

Recent work has increasingly applied MoE architectures to time series forecasting. Mixture-of-Linear-Experts (MoLE) \citep{ni2024mixture} extends linear forecasting models such as DLinear \citep{zeng2023transformers}, RLinear, and RMLP \citep{li2023revisiting} with multiple linear experts and a timestamp-conditioned router, enabling experts to specialize in distinct seasonal or periodic behaviors. LeMoLE \citep{zhang2024lemole} further expands this framework by incorporating LLM-derived features for multimodal forecasting.

At larger scales, TimeMoE \citep{shi2025timemoe} introduces a sparse MoE-based decoder-only foundation model for time series forecasting. Moirai-MoE \citep{liu2024moirai} similarly integrates sparse expert layers into a unified time series foundation model. Together, these studies demonstrate that MoE provides an effective inductive bias for time series forecasting. However, like prior MoE research, they primarily rely on global forecasting objectives and router-side auxiliary losses, without directly supervising expert-level prediction quality. An exception is Mixture-of-Gaussians with
Uncertainty-based Gating (MoGU) framework introduced by \cite{aviv2025mogu}, which incorporates expert-specific uncertainty to determine the gating weights.

Beyond architectural design, recent work has started to ask whether classical MoE training strategies transfer to forecasting. \cite{yemets2025loadbalancing} systematically compare three families of load-balancing strategies for MoE-based time-series forecasting and find that classical auxiliary losses imported from language modeling can actively degrade forecasting accuracy. This motivates the exploration of alternative forms of expert-level supervision, which our expert-specific loss formulation is designed to address.

\section{Model design}

\begin{figure}[h!]
\centering
% You can pre-specify the width of the graph:
\includegraphics[width=0.7\textwidth]{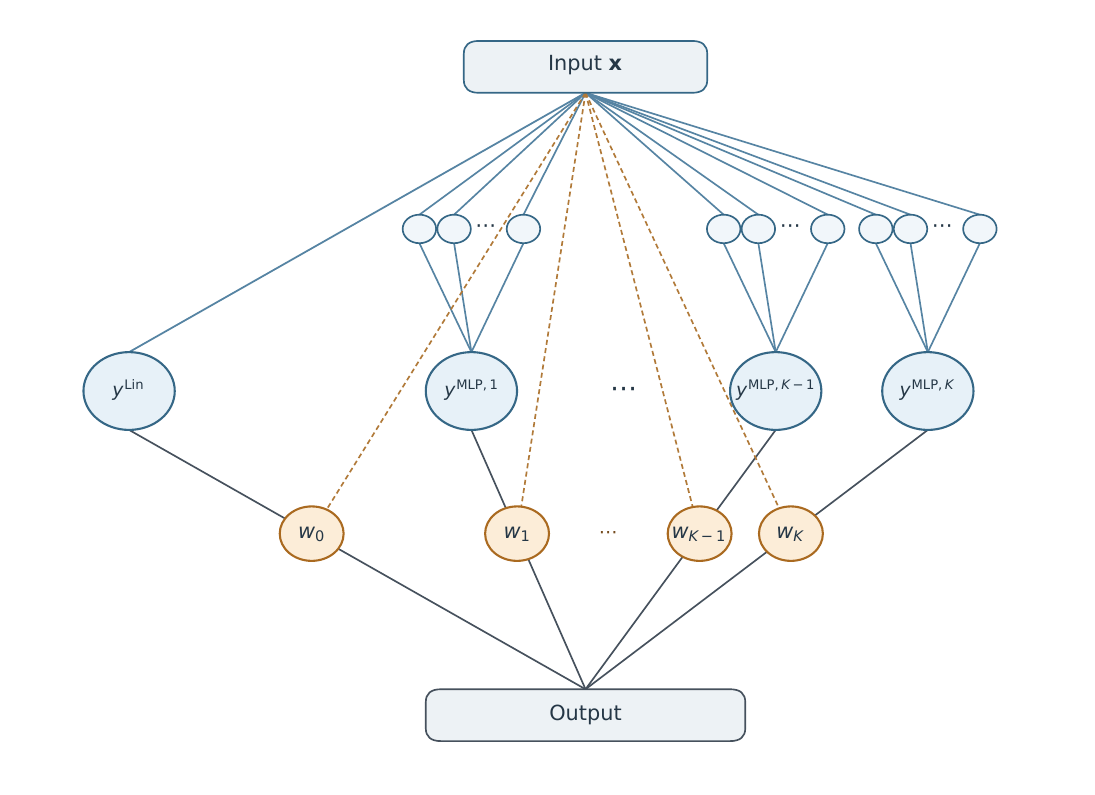}
\caption{\label{fig1} Schematic of EliMoE architecture with heterogeneous expert branches.}
\end{figure}

 We propose a multivariate neural MoE framework which incorporates $K+1$ experts. The first expert is a linear model that directly connects the input to its output, denoted by $Y^{\mathrm{Lin}}(\mathbf{x})$. Each of the remaining $K$ experts is a multilayer perceptron (MLP) with one hidden layer, denoted as $Y^{\mathrm{MLP},k}(\mathbf{x})$.
The architecture of each $\mathrm{MLP}$ may vary, particularly with respect to the number of neurons in the hidden layer. The model is expressed as follows:
\begin{align*}
Y
&=
w_0(\mathbf{x}) \, Y^{\mathrm{Lin}}(\mathbf{x})
+ \sum_{k=1}^K w_k(\mathbf{x}) \, Y^{\mathrm{MLP},k}(\mathbf{x})
+ \xi,
%\label{eq3}
\\
Y^{\mathrm{Lin}}(\mathbf{x})
&=
\mathbf{x}^\top \boldsymbol{\beta},
\qquad
Y^{\mathrm{MLP},k}(\mathbf{x})
=
\mathbf{W}_k^{(2)}\,\sigma\!\left(
\mathbf{W}_k^{(1)}\,\mathbf{x}
+ \mathbf{b}_k^{(1)}
\right)
+ \mathbf{b}_k^{(2)}.
\end{align*}
where 
\(\mathbf{x} \in \mathbb{R}^{m}\) is the input vector with $m$ regressors. $\boldsymbol{\beta}$ is the vector of weights for the skip connection.  
\(\mathbf{W}_k^{(1)}, \mathbf{W}_k^{(2)}, \mathbf{b}_k^{(1)}, \mathbf{b}_k^{(2)}\) 
are the trainable weight matrices and bias vectors, respectively, for k = 1,...,K.  $\sigma$ is the activation function for the hidden layer. %, where we suggest the LeakyReLU:
% $$\sigma(z) = \text{LeakyReLU}(z;\alpha) \;=\;
%\max(z,0)+\alpha \min(z,0),$$
%with $\alpha=0.01$ represents the leakage factor in the LeakyReLU activation function.
 %The $K+1$ mixing weights $\boldsymbol{w} = (w_0,\ldots, w_K)$ 
 $w_j(\mathbf{x})$ denotes the weight assigned to each expert by the gating mechanism and expressed as
 $$w_j(\mathbf{x})
=
\frac{\exp\!\left(g_j(\mathbf{x})\right)}
{\displaystyle\sum\nolimits_{\ell=0}^{K}\exp\!\left(g_\ell(\mathbf{x})\right)},
\qquad
g_j(\mathbf{x})
=
\sigma\!\left(\mathbf{a}_j^\top \mathbf{x}+c_j\right),
\qquad
j=0,\ldots,K,$$

where $\mathbf{a}_j$ denotes the vector of weights connecting the input features to the gating neuron associated with expert $j$ and $c_j$ is its bias term.
Figure \ref{fig1} illustrates the architecture of EliMoE. The blue connections correspond to the expert networks. The orange gating weights are generated from neurons directly connected to the input, followed by a LeakyReLU activation and a softmax function to obtain the normalized expert weights $w_j(\mathbf{x})$.

To estimate the parameters of EliMoE, we introduce a novel objective function that incorporates the individual losses of each expert, as defined below:

% \begin{align}
% \mathcal{L}(\boldsymbol{\theta})
% &= \gamma\,\mathcal{L}_{\mathrm{base}} + (1-\gamma)\,\mathcal{L}_{\mathrm{aux}},
% \label{eq:loss_total}
% \\[0.5em]
% \mathcal{L}_{\mathrm{base}}
% &= \frac{1}{N}\sum_{n=1}^{N}\left|Y_n-\widehat{Y}_n\right|,
% \nonumber
% \\[0.5em]
% \mathcal{L}_{\mathrm{aux}}
% &= \frac{1}{K+1}\left(\frac{1}{N}\sum_{n=1}^{N}\left|Y_n-\widehat{Y}^{\mathrm{Lin}}_n\right|+\sum_{k=1}^{K}\frac{1}{N{\color{red}-\lfloor\pi_k N\rfloor}}\sum_{n=1}^{N}{\color{red}\mathbb{I}\!\left(n>\lfloor\pi_k N\rfloor\right)}\left|Y_n-\widehat{Y}^{\mathrm{MLP},k}_n\right|\right),
% \nonumber
% \\[0.5em]
% &\text{with }{\pi_k}=\frac{k-1}{K},\quad \gamma\in[0,1]\quad \text{and }k=1,\ldots,K.
% \nonumber
% \end{align}

\begin{align}
\mathcal{L}(\boldsymbol{\theta})
&= \gamma\,\mathcal{L}_{\mathrm{base}} + (1-\gamma)\,\mathcal{L}_{\mathrm{aux}},
\label{eq:loss_total}
\\[0.5em]
\mathcal{L}_{\mathrm{base}}
&= \frac{1}{N}\sum_{n=1}^{N}\ell(Y_n,\widehat{Y}_n),
\nonumber
\\[0.5em]
\mathcal{L}_{\mathrm{aux}}
&= \frac{1}{K+1}\left(\frac{1}{N}\sum_{n=1}^{N}\ell(Y_n,\widehat{Y}^{\mathrm{Lin}}_n)
+\sum_{k=1}^{K}\frac{1}{N{\color{red}-\lfloor\pi_k N\rfloor}}
\sum_{n=1}^{N}{\color{red}\mathbb{I}\!\left(n>\lfloor\pi_k N\rfloor\right)}
\ell(Y_n,\widehat{Y}^{\mathrm{MLP},k}_n)\right),
\nonumber
\\[0.5em]
&\text{with }{\pi_k}=\frac{k-1}{K},\quad \gamma\in[0,1]\quad \text{and }k=1,\ldots,K,
\nonumber
\end{align}

where $\ell(\cdot)$ denotes the loss function and $\mathbb{I}(\cdot)$ is the indicator function. Throughout this study, we employ the absolute error ($L_1$) loss, although the proposed framework can accommodate alternative loss functions, such as squared error ($L_2$) or negative log-likelihood.

The objective function in \eqref{eq:loss_total} consists of two components.
The term $\mathcal{L}_{\mathrm{base}}$ represents the overall prediction error of the MoE model. This term is weighted by the parameter $\gamma$, which controls the contribution of the global model prediction. The auxiliary loss $\mathcal{L}_{\mathrm{aux}}$ captures the individual performance of the experts. The first term in $\mathcal{L}_{\mathrm{aux}}$ corresponds to the error of the linear expert. The next $K$ terms correspond to masked losses of the MLP experts.

The proposed objective extends the standard MoE training objective by introducing an auxiliary expert-level loss. This modification is motivated by a specific optimization issue in conventional MoE training.  In particular, the parameters $\theta_j$ of expert $j$ are updated according to the gradient of the overall loss with respect to $\theta_j$. Under the standard mixture objective, this gradient can be written as

 $$
\nabla_{\theta_j}L_{\mathrm{base}}=w_j(\mathbf{x})
\frac{\partial L_{\mathrm{base}}}{\partial\hat{y}}
\cdot
\frac{\partial\hat{y}_j}{\partial\theta_j}.
$$

This expression shows that the learning signal received by expert $j$ is directly scaled by its gating weight. Consequently, when $w_j(\mathbf{x})$ is small, the corresponding expert receives only a weak gradient and may therefore learn slowly or remain undertrained.

Under the proposed objective, the gradient with respect to the parameters $\theta_j$ of expert $k$ is given by

$$
\nabla_{\theta_j}L=\gamma\nabla_{\theta_j}L_{\mathrm{base}}
+
(1-\gamma)\nabla_{\theta_k}L_{\mathrm{aux}}.
$$

Unlike the gradient of the base loss, $\nabla_{\theta_j}L_{\mathrm{aux}}$ is not scaled by the current gating weight. The auxiliary loss therefore provides each expert with a direct supervision signal from the target, including experts that are temporarily assigned low routing probabilities. Hence, including the auxiliary loss addresses the weak-gradient problem caused by small gating weights.

In addition to improving the optimization process, the auxiliary loss in \eqref{eq:loss_total} encourages experts to learn from different temporal segments of the data by applying an expert-specific masking strategy. In particular, the red terms in $\mathcal{L}_{\mathrm{aux}}$ determine the subset of observations used to train each expert, such that different experts are exposed to different portions of the training window, thereby giving each expert a distinct temporal learning task. Experts trained on shorter segments focus more strongly on recent observations and can adapt to short-term changes, whereas experts trained on longer segments use a broader history and can capture more persistent patterns.

%To ensure computational efficiency during training, we adopt the partial online learning approach proposed by El Mahtout and Ziel (2026). This approach combines warm-start initialization with window-specific hyperparameters to reduce the overall training time. In the first training window, the model is initialized randomly. From the second window onward, the model is initialized using the optimal weights obtained from the previous window. Furthermore, distinct hyperparameter configurations are employed for the initial window and subsequent windows, allowing the training procedure to be tailored to different stages of the learning process.

%It is worth highlighting that the proposed objective function is independent of the architecture of the individual experts. In this study, we employ linear and MLP experts, as our primary objective is to develop a model that improves forecasting accuracy without introducing excessive computational overhead. Nevertheless, the proposed MoE framework is sufficiently flexible to accommodate other types of experts. Additionally, under a rolling-window forecasting scheme, a computationally efficient model architecture alone is not sufficient to ensure fast training, since the model must be repeatedly updated after each forecasting step. To address this challenge, we combine the proposed MoE framework with the partial online learning strategy described in Section \ref{4.2}.

It is worth highlighting that the main objective of this work is to achieve a favorable trade-off between forecasting accuracy and computational efficiency. To this end, we adopt two complementary approaches:

\begin{itemize}
    \item Propose a novel objective function that incorporates expert-specific losses to improve forecasting accuracy. Importantly, this objective is independent of the architecture of the individual experts and can therefore be applied to different MoE frameworks.
    
    \item  Employ computationally efficient experts, namely linear and MLP models, to avoid excessive computational overhead. However, under a rolling-window forecasting scheme, a computationally efficient architecture alone is insufficient to ensure fast training, as the model must be repeatedly updated after each forecasting step. To address this challenge, we combine the proposed MoE framework with the partial online learning strategy described in Section \ref{4.2}.
\end{itemize}

\section{Experiments}\label{Experiments}
\subsection{Fixed-Scheme Forecasting Study}
 We conduct a fixed-scheme forecasting study using five datasets from diverse fields with varying frequencies. These datasets are extracted from Monash Time Series Forecasting Repository \citep{godahewa2021monash}. The first dataset is the electricity demand of 321 individuals  originally sourced from the UCI repository \citep{UCI_electricity_load}. We use the aggregated weekly series for our analysis. %The Dominick dataset is another weekly dataset that tracks retail operations and contains 115704 time series, sourced from the University of Chicago Booth School of Business's online platform \cite{Dominick}. 
 In addition to this energy dataset, we also include the hourly M4 competition dataset \citep{makridakis2018m4, makridakis2020m4}, which comprises 100,000 time series. Furthermore, we use the monthly tourism dataset, consisting of 1,311 tourism-related time series originating from a Kaggle competition \citep{athanasopoulos2011tourism, ellis2018tcomp}, as well as a monthly car parts dataset comprising 2,674 time series representing sales data. The latter dataset is originally provided in the expsmooth R package \citep{hyndman2015expsmooth} and contains missing values. In this study, we use the version where missing values have been imputed. The final dataset is the Saugeen dataset, which consists of a single long time series with 23,741 daily observations of the mean flow of the Saugeen River. This dataset is sourced from the deseasonalize R package \citep{mcleod2013optimal}.

%For the Saugeen River dataset, we employ an MoE architecture with four experts: one linear expert and three MLP experts. For the remaining datasets, however, we use a simpler MoE structure consisting of two experts: one linear expert and one MLP expert without masking. This design choice is motivated by the nature of the datasets. Unlike the Saugeen River dataset, the remaining datasets contain multiple distinct time series. Consequently, applying masking based on consecutive observations may be misleading, since adjacent observations can originate from different series. For further details regarding the hyperparameters and forecast horizons, refer to Appendix \ref{forecast_study}.

To generate forecasts for each of the datasets described above, we employ EliMoE comprising four experts: one linear expert and three MLP experts. The model uses only lagged values of the target series as regressors. For each dataset, the number of lags and the forecast horizon are set according to the specifications used in \cite{godahewa2021monash}. The exact values are reported in Table \ref{lags_hotizon} in the appendix. The forecasting accuracy and computational efficiency of EliMoE are then compared with those of several benchmark models, including the feed-forward neural network \citep[FFNN,][]{Goodfellow_2016}, DeepAR \citep{salinas2020deepar}, N-BEATS \citep{nbeats}, WaveNet \citep{borovykh2017conditional}, the Transformer \citep{vaswani2017attention}, PatchTST \citep{nie2022time}. In addition, we consider four MoE-based benchmarks.  MoLE with an RMLP backbone \citep{li2023revisiting,ni2024mixture}, MoGU \citep{aviv2025mogu} with DLinear \citep{zeng2023transformers}, and two zero-shot time-series foundation models: TimeMoE  \citep{shi2025timemoe} and  Moirai-MoE \citep{liu2024moirai}.%In addition, we consider three MoE-based benchmarks: TimeMoE, a zero-shot time-series foundation model \cite{shi2025timemoe}, and Mixture-of-Linear-Experts (MoLE) with an RMLP backbone \cite{li2023revisiting,ni2024mixture}.

The forecast comparison between the proposed model and the benchmark models is based on two popular error metrics: Mean Absolute Error \citep[MAE,][]{mae2010} and Root Mean Squared Error (RMSE). In addition, Mean Absolute Scaled Error \citep[MASE,][]{hyndman2006another} is considered, as it provides scale-independent evaluation and accounts for seasonality when a seasonal naïve benchmark is used. Since most of the dataset contains multiple time series, %we report both the mean and median values of each evaluation metric to summarize overall performance.
we report mean values of each evaluation metric to summarize overall performance.

%Table \ref{tab:mase_runtime} presents the MASE values for each model across the datasets, along with the computational time required to fit the models and generate forecasts. Among the univariate statistical models, TBATS and ARIMA generally perform better than SES, Theta, and ETS.  In particular, TBATS achieves relatively strong results on Electricity Weekly (0.79) and Saugeen River Flow (1.43), but its performance deteriorates on the Carparts dataset.

To ensure a rigorous experimental setup, we reran EliMoE and all baseline models 10 times using different random seeds and report the mean and standard deviation of the respective performance metrics and computational runtime. Table \ref{tab:mase_runtime} reports the mean MASE across the 10 runs, together with the average computational time required for model fitting and forecast generation.

\begin{table}[h!]
\centering
\caption{Comparison of forecasting accuracy and computational cost across datasets and models. Each model is represented by two rows: the first reports the mean MASE, and the second reports the mean runtime in seconds, with the corresponding standard deviations shown in parentheses. For each dataset, the lowest mean MASE and the fastest mean runtime are highlighted in bold. The smallest standard deviations for MASE and runtime are underlined across all models.}
\label{tab:mase_runtime}
\resizebox{\textwidth}{!}{%
\begin{tabular}{lrrrrr}
\toprule
\textbf{Model} & \textbf{Electricity weekly} & \textbf{M4 hourly} & \textbf{Saugeen day} & \textbf{Tourism monthly} & \textbf{Car parts} \\
\midrule
\textbf{EliMoE} & \textbf{0.76} (\underline{0.01}) & 1.47 (0.05) & \textbf{1.40} (\underline{0.01}) & 1.49 (\underline{0.01}) & \textbf{0.75} (\underline{0.00}) \\
 & \textbf{3.08} (0.51) & 35.52 (\underline{0.79}) & \textbf{2.14} (\underline{0.12}) & \textbf{8.98} (\underline{0.65}) & 3.37 (0.30) \\
\midrule
\textbf{MoGU} & 0.87 (\underline{0.01}) & \textbf{0.77} (\underline{0.00}) & 1.51 (0.03) & 1.55 (0.02) & 1.14 (0.02) \\
 & 23.51 (3.68) & 604.00 (46.99) & 17.97 (0.35) & 85.67 (36.10) & \textbf{2.59} (\underline{0.14}) \\
\midrule
\textbf{MoLE} & 0.82 (0.06) & 3.02 (0.67) & 1.55 (0.05) & 1.63 (0.08) & 0.90 (0.01) \\
 & 8.80 (1.49) & 142.56 (26.71) & 5.27 (0.35) & 26.95 (7.69) & 2.91 (0.30) \\
\midrule
\textbf{Moirai-MoE} & 1.44 (0.00) & 1.01 (0.00) & 1.96 (0.05) & 3.11 (0.01) & 0.77 (0.00) \\
 & 1.25\textsuperscript{*} (0.07) & 109.14\textsuperscript{*} (1.87) & 0.55\textsuperscript{*} (0.30) & 13.10\textsuperscript{*} (0.45) & 3.62\textsuperscript{*} (0.06) \\
\midrule
\textbf{TimeMoE} & 1.09 & 0.88 & 2.13 & 2.88 & 1.07 \\
 & 4.85\textsuperscript{*} & 22.54\textsuperscript{*} & 0.98\textsuperscript{*} & 4.28\textsuperscript{*} & 27.54\textsuperscript{*} \\
\midrule
\textbf{PatchTST} & 0.90 (0.04) & 3.11 (0.16) & 1.58 (0.02) & 1.77 (0.03) & 0.93 (0.01) \\
 & 274.83 (5.59) & 9560.95 (28.53) & 96.73 (0.97) & 409.28 (17.26) & 159.94 (4.09) \\
\midrule
\textbf{DeepAR} & 1.17 (0.12) & 2.57 (0.76) & 1.92 (0.20) & \textbf{1.44} (0.03) & \textbf{0.75} (\underline{0.00}) \\
 & 334.51 (19.37) & 1302.47 (21.64) & 255.93 (12.36) & 179.53 (16.53) & 169.70 (2.15) \\
\midrule
\textbf{FFNN} & 0.80 (0.05) & 2.67 (0.61) & 1.42 (0.04) & 1.61 (0.03) & \textbf{0.75} (\underline{0.00}) \\
 & 15.71 (1.03) & \textbf{23.61} (1.24) & 40.42 (4.60) & 14.73 (1.45) & 16.64 (0.29) \\
\midrule
\textbf{N-BEATS} & 0.78 (0.04) & 1.91 (0.34) & 1.87 (0.08) & 1.57 (0.03) & 2.84 (\underline{0.00}) \\
 & 1245.82 (85.93) & 1443.99 (5.14) & 1203.38 (34.46) & 1302.66 (12.49) & 1491.96 (147.88) \\
\midrule
\textbf{Transformer} & 1.33 (0.15) & 4.87 (1.82) & 1.87 (0.06) & 1.79 (0.48) & \textbf{0.75} (\underline{0.00}) \\
 & 277.13 (\underline{0.47}) & 1620.51 (15.33) & 212.45 (1.80) & 146.91 (7.50) & 150.79 (0.63) \\
\midrule
\textbf{WaveNet} & 1.11 (0.09) & 1.80 (0.20) & 1.52 (0.06) & 1.49 (0.03) & 0.76 (0.02) \\
 & 613.13 (51.49) & 2939.62 (283.52) & 946.30 (55.27) & 696.03 (51.21) & 334.39 (29.45) \\
\bottomrule
\end{tabular}
}
\par\smallskip
\begin{minipage}{\textwidth}
\footnotesize\raggedright
\textsuperscript{*}TimeMoE and Moirai-MoE runtime corresponds to data loading and forecast generation only.
\end{minipage}
\end{table}

The results in Table \ref{tab:mase_runtime} show that the EliMoE achieves strong forecasting performance across the considered datasets. It attains the lowest MASE on the Electricity Weekly and Saugeen River flow datasets. On the Carparts dataset, its performance is comparable to that of the best-performing models, including FFNN, DeepAR, and Transformer. Although EliMoE does not achieve the lowest MASE on the Tourism Monthly and M4 Hourly datasets, it ranks among the three best-performing models on both datasets, demonstrating its competitiveness across diverse forecasting tasks. The superiority of the proposed model remains evident when alternative error metrics are considered, as shown in Tables \ref{tab:mean_mae} and \ref{tab:mean_rmse} in Appendix \ref{tables}. %Although EliMoE does not achieve the best performance on every dataset under all evaluation metrics, it consistently exhibits competitive performance, placing it among the top-performing models overall.

In addition to its strong forecasting performance, EliMoE achieves the lowest computational runtime on three datasets and ranks among the most computationally efficient models on the remaining two. Its computational advantage over the other benchmark models is often substantial. For instance, on the Electricity weekly dataset, EliMoE requires approximately three seconds for model training and forecast generation, whereas the Transformer requires approximately 90 times as much time. Another noteworthy result in Table \ref{tab:mase_runtime} is the low variability of EliMoE across different random seeds. Across the considered datasets, EliMoE generally achieves the smallest standard deviation for both MASE and computational time, suggesting that its forecasting accuracy and runtime are relatively stable across repeated runs.

It is important to note that TimeMoE and Moirai-MoE are excluded when identifying the fastest model, as they are pretrained models whose reported runtimes in Table \ref{tab:mase_runtime} include only model loading and forecast generation. Additionally, the standard deviations reported for Moirai-MoE reflect variability arising from stochastic inference rather than random initialization and training. Therefore, Moirai-MoE is excluded when identifying the smallest standard deviations across models.

\subsection{Online learning}\label{4.2}

The forecasting study in the previous section employed the fixed scheme, in which the model is trained once. The same weights are used to generate forecasts  over the entire horizon. This approach is computationally inexpensive but often comes at the cost of reduced accuracy, particularly for longer forecast horizons. In time series analysis, there are other forecasting approaches that yield more accurate forecasts, such as the rolling window scheme. In this approach, the model is repeatedly retrained using a moving training window. However, this repeated retraining is computationally very expensive, particularly since each subsequent training window differs from the previous one by only a single additional observation. %Specifically, for a window size \(D\), the model is first trained on the \(D\) observations immediately preceding the first point in the test set, and a forecast is generated for that point. The window then shifts forward by one observation: the model is re-trained on the next \(D\) observations (ending just before the second test point), and a forecast is produced for the second point. This process continues sequentially through the test set. 

%As a result, the model is re-trained \(P\) times, where \(P\) is the length of the test set. Since the forecast is generated one step ahead for each training window across the entire test set, \(P\) is equivalent to the forecast horizon \(h \) in the fixed-scheme forecast. However, this repeated retraining is computationally very expensive, particularly since each subsequent training window differs from the previous one by only a single additional observation. 

To minimize the computational burden introduced by the rolling window scheme, we combine the proposed MoE with partial online learning introduced by \cite{mahtout2026electricity}. This approach reduces training time through two key mechanisms. First, it employs warm starting, where model weights from the previous training window are used to initialize the current window, thereby accelerating convergence. Second, it uses two distinct sets of hyperparameters: one optimized for the initial training window and another shared across all subsequent windows. 

%In the first training window, model weights are initialized randomly, and a set of hyperparameters specifically tailored to this stage is used. For example, a larger window size and a higher number of training epochs are employed to allow the model to effectively learn the underlying structure of the time series. In the second training window, the model is initialized using the optimized weights obtained from the first window. At this stage, a different set of hyperparameters is applied, typically involving a smaller window size and fewer training epochs. This adjustment is justified because the model already captures relevant patterns from the previous window, and only a single new observation differentiates the current window from the prior one. From the third training window onward, up to the final window, the model continues to use the weights from the immediately preceding window as initialization. The same hyperparameter configuration introduced in the second window is retained throughout these subsequent training steps, ensuring computational efficiency while maintaining model performance.

We conduct a rolling-window forecasting study on the Saugeen River Flow dataset, combining EliMoE with the partial online learning strategy. The forecasting accuracy and execution time of the resulting model are then compared with those of the FFNN, MoLE, and MoGU benchmarks. Since none of these baseline models explicitly incorporates an online learning mechanism, we include an additional benchmark specifically designed for online forecasting to ensure a more balanced comparison. Specifically, we employ ADAPT-Z \citep{huang2026online} with an iTransformer backbone \citep{liu2024itransformer}. 

\begin{figure}[h!]
\centering
% You can pre-specify the width of the graph:
\includegraphics[width=0.8\textwidth]{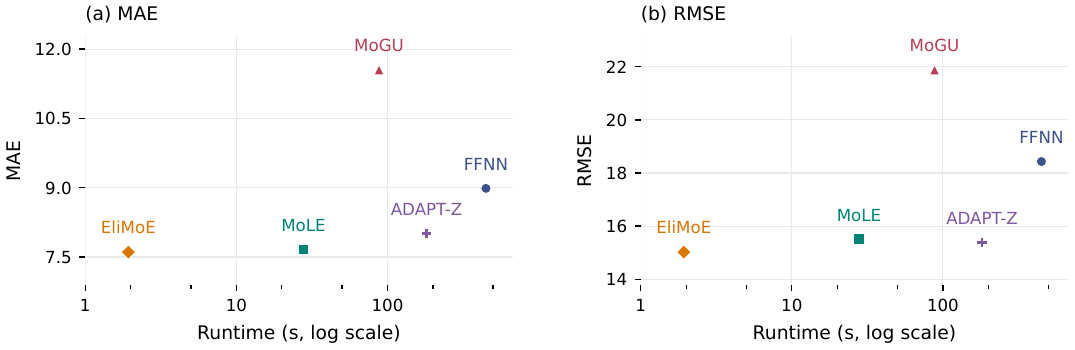}
\caption{Forecasting accuracy and computational efficiency of EliMoE with partial online and four benchmark models under the rolling-window forecasting scheme on the Saugeen Day dataset. %Each marker represents the mean forecasting error and runtime across 10 random seeds. 
}
\label{fig2}
\end{figure}

EliMoE and other benchmark models are hyperparameter-tuned using the Tree-structured Parzen Estimator (TPE) optimization method \citep{TPE}, implemented through the Python package Optuna \citep{optuna}. Further details on the selected hyperparameters and the validation dataset are provided in Appendix \ref{forecast_study}. During hyperparameter tuning, the best-performing set of hyperparameters is selected for each model and subsequently used to generate forecasts on the test data across ten different random seeds. Figure \ref{fig2} illustrates the average MAE, RMSE, and computational time across the ten seeds for the considered models. As expected, all models demonstrate substantially improved performance compared with the fixed-scheme framework. For example, the MAE of the FFNN decreases from 21.39 under the fixed scheme (see Table \ref{tab:mean_mae}  in the Appendix) to 8.98 under the rolling window strategy. Similar improvements are also observed for other models.

%Table \ref{tab:moe_ffnn_comparison} presents a comparison between the MoE model with online learning and the FFNN under the rolling window forecasting framework. As expected, both MoE and FFNN demonstrate substantially improved performance compared with the fixed-scheme framework. For example, the MAE of the FFNN decreases from 22.83 under the fixed scheme (see Table \ref{tab:median_mae} or  in the Appendix) to 8.11 under the rolling window strategy. Similar improvements are also observed for the MoE model.

%Within the rolling-window framework, EliMoE clearly outperforms the FFNN, reducing MAE by 17\% and RMSE by 18\%. Compared with MoLE, EliMoE achieves an 11\% reduction in MAE, while RMSE increases marginally, by less than 1\%. EliMoE also demonstrates substantially higher computational efficiency, requiring approximately two seconds to complete the full forecasting study, compared with approximately eight minutes for the FFNN and one minute for MoLE. Additionally, the error bars indicate that EliMoE exhibits lower variability across the ten random seeds than the other models under the rolling-window scheme, which is consistent with the results observed under the fixed forecasting scheme.

Within the rolling-window forecasting framework, EliMoE outperforms all four baseline models, achieving reductions in RMSE ranging from 2\% to 31\% and reductions in MAE ranging from less than 1\% to 34\%. EliMoE also demonstrates substantially higher computational efficiency, requiring approximately two seconds to complete the entire forecasting experiment, compared with approximately 28 seconds for MoLE, the second-fastest model. Notably, EliMoE achieves a shorter runtime even compared with ADAPT-Z, which explicitly incorporates an online learning mechanism. These results highlight EliMoE’s ability to achieve superior forecasting accuracy while maintaining low computational costs under the considered experimental setting.

%Additionally, the error bars indicate that EliMoE exhibits lower variability across the ten random seeds than the other models under the rolling-window scheme, which is consistent with the results observed under the fixed forecasting scheme.

%It is worth noting that the computational time of the MoE with online learning, despite being refitted at each window, is approximately the same as that of the fixed-scheme approach, which is fitted only once.

\subsection{Ablation study}

%The parameter \(\gamma\) in the loss function defined in Equation \ref{eq:loss_total} is a crucial component of the proposed methodology. It takes values in the interval [0,1]. When \(\gamma= 0\), only the expert-specific losses contribute to the training process, while the aggregated forecasting loss is ignored. In contrast, when \(\gamma= 1\), the expert-specific losses no longer participate in training, and the model reduces to the standard formulation based solely on the aggregated forecasting loss. Note that throughout this study, we set \(\gamma=0.25 \), thereby assigning greater importance to expert-specific losses than to the aggregated forecasting loss.

To assess the effectiveness of the proposed framework, we conduct four experiments designed to isolate the contributions of its main components. Specifically,
the effect of the weighting coefficient $\gamma$, the impact of the masking strategy, we investigate the role of the auxiliary loss in mitigating weak gradient signals for experts with low gating weights, the trade-off between forecasting accuracy and computational cost as the number of experts varies, the generalizability of the proposed framework, and the individual contributions of the lightweight expert architecture and partial online learning strategy to computational efficiency. 
All experiments are conducted using the Saugeen River Flow dataset under the rolling-window forecasting scheme, except for the analysis of the impact of the number of experts, which is conducted under the fixed forecasting scheme.

% To investigate how the proposed objective affects expert training under weak gating signals, we initialize the gating parameters of three experts to zero and assign a negative initial value to the fourth expert to induce a very small gating weight. We then monitor the gradient norm of the fourth expert after each parameter update, together with its MAE evaluated over the entire dataset. Figure \ref{mechnistic_plot} illustrates the results
%  \begin{figure}[h!]
% \centering
% % You can pre-specify the width of the graph:
% \includegraphics[width=0.8\textwidth]{mechanistic_gradients_mae.pdf}
% \caption{Effect of auxiliary loss on expert learning under weak 
% gating signals. The figure illustrates the evolution of the gradient norm and expert-specific MAE evaluated 
% over the entire dataset across successive optimization updates.}
% \label{mechnistic_plot}
% \end{figure}

% The left panel of Figure \ref{mechnistic_plot} demonstrates that the gradient norm remains very small when $\gamma=1$, whereas substantially larger gradient norms are observed when $\gamma=0.5$. This illustrates the direct effect of incorporating the auxiliary loss on the gradient signal and, consequently, on the parameter updates of the corresponding expert. The right panel further demonstrates that incorporating the auxiliary loss improves the forecasting accuracy of the expert in the case of a weak gating weight. These findings provide empirical evidence that the proposed objective mitigates gradient attenuation caused by weak gating weights.

The parameter $\gamma$ in the objective function defined in Equation \ref{eq:loss_total} is a crucial component of the proposed methodology. It takes values in the interval [0,1]. When $\gamma= 0$, only the expert-specific losses contribute to the training process, while the aggregated forecasting loss is ignored. In contrast, when $\gamma= 1$, the expert-specific losses no longer participate in training, and the model reduces to the standard formulation based solely on the base loss. Note that throughout this study, we set $\gamma=0.5$, thereby assigning equal weight to the base loss and the expert-specific losses.

% The first two rows of Table \ref{tab:moe_gamma_masking} illustrate the impact of different values of $\gamma$ on the forecasting accuracy of EliMoE with temporal masking specified in the auxiliary loss $\mathcal{L}_{\mathrm{aux}}$ . For $\gamma \in \{0.25, 0.5, 0.75\}$, MAE decreases by approximately 5\% relative to $\gamma=1$, which corresponds to the standard MoE setting. Similarly, RMSE improves by approximately 4\%. These results indicate that incorporating expert-specific losses together with the masking strategy can improve forecasting accuracy compared with relying solely on the base loss.

The first two masking strategies reported in Table \ref{tab:moe_gamma_masking} allow us to examine how the choice of $\gamma \in \{0.25, 0.5, 0.75\}$ interacts with the temporal allocation of training data across experts. The first strategy follows the nested temporal masking scheme defined in the auxiliary loss $\mathcal{L}_{\mathrm{aux}}$. In the EliMoE specification used in this study, the model consists of one linear expert and three MLP experts. When computing the auxiliary expert-specific losses, the linear expert and $\mathrm{MLP}^{1}$ use the full training window, whereas $\mathrm{MLP}^{2}$ uses the most recent two-thirds of the observations and $\mathrm{MLP}^{3}$ uses only the most recent one-third. This nested partitioning ensures that all experts have access to the most recent observations. Note that, when computing the base loss, all experts have access to the full training window. In this nested masking setup,  the best MAE and RMSE are reduced by approximately 5\% and 4\%, respectively, compared with the standard MoE formulation ($\gamma=1$).

% \begin{table*}[h!]
% \centering
% \captionsetup{skip=4pt}
% \caption{Effect of $\gamma$ and masking strategy (temporal masking, disjoint masking, and without masking) on EliMoE online forecasting
% accuracy on the Saugeen day dataset. Values are means over 10 seeds. Bold indicates the lowest error in each row}
% \label{tab:moe_gamma_masking}
% \scriptsize
% \setlength{\tabcolsep}{5pt}
% \renewcommand{\arraystretch}{1.1}
% \begin{tabular*}{\textwidth}{@{\extracolsep{\fill}}llrrrrr@{}}
% \toprule
% \textbf{Setting} & \textbf{Metric} & $\boldsymbol{\gamma}=0$ & $\boldsymbol{\gamma}=0.25$ & $\boldsymbol{\gamma}=0.5$ & $\boldsymbol{\gamma}=0.75$ & $\boldsymbol{\gamma}=1$ \\
% \midrule
% \multirow{2}{*}{Temporal masking} & MAE & 9.40 & 8.42 & \textbf{7.61} & 8.05 & 7.98 \\
%  & RMSE & 17.23 & 15.81 & \textbf{15.02} & 16.99 & 15.63 \\
% \midrule
% \multirow{2}{*}{Disjoint masking} & MAE & 10.09 & 8.05 & 8.12 & \textbf{7.59} & 7.98 \\
%  & RMSE & 19.14 & 15.52 & 16.11 & \textbf{14.89} & 15.63 \\
% \bottomrule
% \multirow{2}{*}{Without masking} & MAE & \textbf{7.77} & 7.96 & 8.48 & 8.44 & 7.98 \\
%  & RMSE & 15.93 & 16.08 & 17.04 & 17.01 & \textbf{15.63} \\
% \midrule
% \end{tabular*}
% \end{table*}

The alternative approach is the disjoint temporal masking strategy, in which the training data are partitioned into non-overlapping temporal subsets, each assigned to a different MLP expert through the auxiliary loss. In our architecture, the training data are divided into three equally sized chronological subsets. Each expert receives direct supervision from its respective subset, with only the last expert being exposed to the most recent observations. Unlike nested temporal masking, this strategy therefore limits the extent to which recent information is shared across experts. Despite this potential limitation, disjoint temporal masking also improves forecasting accuracy, yielding reductions of approximately 5\% in both MAE and RMSE relative to the standard MoE formulation. These results suggest that the benefits of incorporating the auxiliary loss are not restricted to a particular temporal masking strategy.

%After examining the impact of different masking designs, we next investigate the importance of the masking mechanism itself. To isolate its contribution to forecasting performance, we retain the auxiliary expert-specific losses in the training objective while removing the masking mechanism. In this setting, all experts use the full training window when computing their auxiliary losses. The last rows of Table \ref{tab:moe_gamma_masking} report the results obtained without masking for different values of $\gamma$. Comparing these results with their corresponding masked configurations shows a clear deterioration in forecasting accuracy when masking is omitted for $\gamma \in \{0.25, 0.5, 0.75\}$. This finding provides  evidence that the masking mechanism contributes to the forecasting performance of EliMoE. An exception is observed for $\gamma=0$, where forecasting performance improves when masking is removed. In this setting, the base loss is completely omitted, and training is driven exclusively by the auxiliary expert-specific losses.

After examining the impact of different masking designs, we next investigate the importance of the masking mechanism itself. To isolate its contribution to forecasting performance, we retain the auxiliary expert-specific losses in the training objective while removing the masking mechanism. In this setting, all experts use the full training window when computing their auxiliary losses. The last rows of Table \ref{tab:moe_gamma_masking} report the results obtained without masking for different values of $\gamma$. Comparing these results with their corresponding masked configurations shows a deterioration in forecasting accuracy when masking is omitted for $\gamma \in \{0.5, 0.75\}$, whereas improvements are observed
for the remaining values of $\gamma$. These findings provide
empirical evidence that both the masking mechanism and the
choice of $\gamma$ play an important role in the effectiveness
of the proposed training objective. In particular, they
highlight the interplay between temporal masking and the
relative weighting of the base and auxiliary expert-specific
losses.

\begin{figure}[h!]
\centering
% You can pre-specify the width of the graph:
\includegraphics[width=1\textwidth,height=4cm]{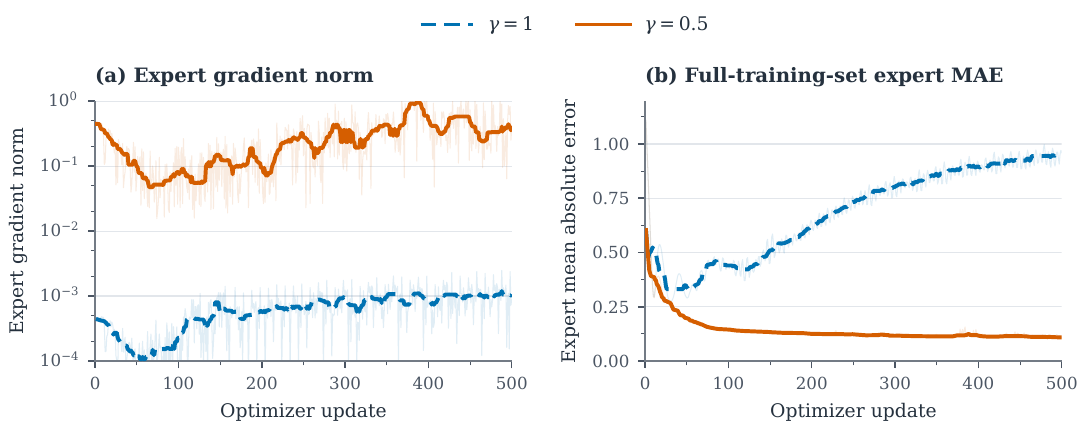}
\caption{Effect of auxiliary loss on expert learning under weak 
gating signals. The figure illustrates the evolution of the gradient norm and expert-specific MAE evaluated 
over the entire dataset across successive optimization updates.}
\label{mechnistic_plot}
\end{figure}

The preceding analysis provides empirical evidence that incorporating expert-specific losses into the training objective generally improves forecasting accuracy. However, it does not provide direct mechanistic evidence of how these auxiliary losses facilitate expert learning under weak gating signals. To address this, we conduct a controlled experiment in which the gating parameters of three experts are initialized to zero, and we assign a negative initial value to the fourth expert to induce a very small gating weight. We then monitor the gradient norm of the fourth expert after each parameter update, together with its MAE evaluated over the entire dataset. Figure \ref{mechnistic_plot} illustrates the results

The left panel of Figure \ref{mechnistic_plot} demonstrates that the gradient norm remains very small when $\gamma=1$, whereas substantially larger gradient norms are observed when $\gamma=0.5$. This illustrates the direct effect of incorporating the auxiliary loss on the gradient signal and, consequently, on the parameter updates of the corresponding expert. The right panel further demonstrates that incorporating the auxiliary loss improves the forecasting accuracy of the expert in the case of a weak gating weight. These findings provide empirical evidence that the proposed objective mitigates gradient attenuation caused by weak gating weights.

The number of experts is another crucial component of the proposed method, particularly when computational efficiency is taken into consideration. To investigate the trade-off between forecasting accuracy and computational cost, we evaluate EliMoE with different numbers of MLP experts and report the corresponding MAE and runtime. Figure \ref{Num_expert_time} in the Appendix summarizes the results. When the number of MLP experts is zero, EliMoE consists solely of the linear expert, resulting in a relatively high MAE but a very short runtime. As the number of MLP experts increases, MAE generally decreases, while computational time increases. The lowest MAE is achieved with eight MLP experts and one linear expert. Beyond this point, increasing the number of experts does not yield further improvements in forecasting accuracy, while computational cost continues to increase.

Finally, to assess whether the proposed methodology generalizes to other MoE architectures, we incorporate the auxiliary expert-specific losses into the training objectives of MoLE and MoGU. The original objective function of MoLE is given by $\frac{1}{N}\sum_{i=1}^{N}\ell\left(y_i,\sum_{k=1}^{K}w_k(\mathbf{x}_i)f_k(\mathbf{x}_i)\right)$, which corresponds to the base loss of our proposed framework when $\ell(\cdot)$ is specified as the squared error loss. Accordingly, we extend the MoLE objective by incorporating the same auxiliary expert-specific loss formulation, with the squared error as the individual expert loss. In contrast, the original objective function of MoGU is given by $\frac{1}{N}\sum_{i=1}^{N}\sum_{k=1}^{K}w_k(\mathbf{x}_i)\mathcal{L}_{\mathrm{NLL}}\left(y_i;\mu_k(\mathbf{x}_i),\sigma_k^2(\mathbf{x}_i)\right)$ denotes the Gaussian negative log-likelihood. To remain consistent with the probabilistic formulation of MoGU, we employ the same negative log-likelihood function when computing the auxiliary expert-specific losses and use the mean of the predictive mixture distribution as the point forecast.

\begin{figure}[h!]
\centering
% You can pre-specify the width of the graph:
\includegraphics[width=0.7\textwidth]{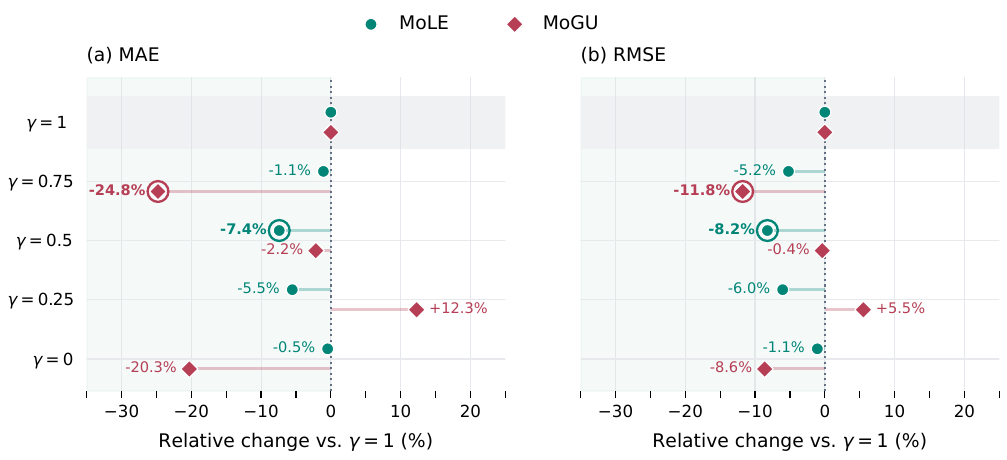}
\caption{Effect of auxiliary expert-specific loss on forecasting accuracy. Relative changes in mean MAE (a) and RMSE (b) for MoLE and MoGU on the Saugeen Day dataset, averaged across 10 random seeds. Each model is compared against its corresponding baseline (\(\gamma=1\)). Negative values indicate improvements in forecasting accuracy, while circled markers highlight the largest relative improvement for each model and metric}
\label{MoLE_MoGU}
\end{figure}

Figure \ref{MoLE_MoGU} illustrates the percentage changes in forecasting error for different values of $\gamma$ relative to the original implementations ($\gamma=1$). With the exception of $\gamma=0.25$, incorporating the auxiliary expert-specific losses consistently improves the forecasting accuracy of both models, with the reduction in MAE reaching approximately 25\% for MoGU. These results provide evidence that the proposed expert-specific loss is not limited to EliMoE and can also improve the forecasting performance of other MoE architectures.

While the preceding analysis investigates the factors contributing to the improvements in forecasting accuracy achieved by the proposed methodology, we now turn our attention to its computational efficiency. Specifically, Figure \ref{Online_tranformer_no_online} in the Appendix illustrates the individual contributions of the two factors previously identified as the main drivers of computational efficiency: the model architecture and the online learning strategy.

In panel (a) of Figure \ref{Online_tranformer_no_online}, we compare the proposed EliMoE architecture, consisting of a linear expert and MLP experts, with an alternative architecture in which all four experts are Transformers. Replacing the original experts with Transformer-based experts substantially increases computational time, resulting in a runtime approximately 19 times that of the original implementation. In panel (b), we retain the same EliMoE architecture employed throughout this study and compare two settings: with and without partial online learning. %The results demonstrate that incorporating partial online learning further reduces computational time. Taken together, these findings highlight the importance of both the lightweight expert architecture and the partial online learning strategy in achieving the computational efficiency of EliMoE.
The results demonstrate that incorporating partial online learning further reduces computational time. Note that although the runtime without partial online learning (13.79 seconds) is substantially higher than that with partial online learning, it remains lower than that of all benchmark models presented in Figure \ref{fig2}, including the online learning model ADAPT-Z. Specifically, the fastest benchmark, MoLE, requires 27.82 seconds. Taken together, these findings highlight the complementary contributions of the lightweight expert architecture and the partial online learning strategy to the computational efficiency of EliMoE.

% \begin{figure}[h!]
% \centering
% % You can pre-specify the width of the graph:
% \includegraphics[width=0.7\textwidth]{rolling_moe_runtime_comparison.pdf}
% \caption{Impact of expert architecture and partial online learning on the computational efficiency of EliMoE on the Saugeen dataset with $\gamma=0.5$.}
% \label{Online_tranformer_no_online}
% \end{figure}

\section{Conclusion}

In this work, we introduced a novel MoE framework for time series forecasting aimed at improving forecasting accuracy while reducing computational cost. The proposed EliMoE incorporates expert-specific losses into the objective function in addition to the aggregated forecasting loss, enabling expert-level prediction errors to directly influence the training process. Additionally, masking is applied to the MLP-specific loss to encourage the network to specialize in a specific subset of the data.

EliMoE demonstrates strong forecasting performance across diverse time series domains, outperforming a broad set of established and state-of-the-art neural forecasting models, including DeepAR, N-BEATS, WaveNet, PatchTST, TimeMoE, and MoLE. The model also achieves substantially higher computational efficiency than the baseline methods. In addition, the proposed methodology is flexible and can be incorporated into other MoE architectures. Empirical results with MoLE and MoGU show that integrating the proposed expert-specific loss improves forecasting accuracy, providing further evidence of the broader applicability of the methodology.

Finally, it is important to acknowledge several limitations of the present study. First, only five values of the $\gamma$ parameter in the objective function were considered, leaving the sensitivity of the proposed objective to a finer range of weighting choices unexplored. In addition, the nested masking strategy ensures that all experts retain access to the most recent observations and has been shown empirically to improve forecasting accuracy. However, there is no guarantee that this masking design is optimal. Future research should therefore investigate the effect of $\gamma$ more comprehensively and explore alternative masking strategies, including learnable temporal partitions.
% \begin{ack}
% This research was partially funded in the course of TRR 391 Spatio-temporal Statistics for the Transition of Energy and Transport (520388526) by the Deutsche Forschungsgemeinschaft (DFG, German Research Foundation) 
% \end{ack}

%%%%%%%%%%%%%%%%%%%%%%%%%%%%%%%%%%%%%%%%%%%%%%%%%%%%%%%%%%%%

% \clearpage

% \bibliography{references}

% \clearpage

% \appendix

% % \section{Appendix}
% \appendixpage
\clearpage
\section*{Declaration on AI-Assisted Tools}

During the preparation of this work, the authors used generative AI tools to improve the readability and clarity of the manuscript, assist with code editing and debugging, and support the literature review.  All AI-assisted content and code were reviewed, verified, and edited by the authors as needed. The authors take full responsibility for the final content of the manuscript.

\begin{ack}
This research was partially funded in the course of TRR 391 Spatio-temporal Statistics for the Transition of Energy and Transport (520388526) by the Deutsche Forschungsgemeinschaft (DFG, German Research Foundation) 
\end{ack}

% \clearpage
% \bibliography{iclr2027_conference}
% \bibliographystyle{iclr2027_conference}

% \appendix
% \section{Appendix}

% % \startcontents[sections]
% % \printcontents[sections]{l}{1}{\setcounter{tocdepth}{2}}
% % \clearpage

\clearpage

\bibliography{references}

\clearpage

\appendix

% \section{Appendix}
\appendixpage

\startcontents[sections]
\printcontents[sections]{l}{1}{\setcounter{tocdepth}{2}}
\clearpage
\section{Tables}\label{tables}

\section{Tables and Figures}\label{tables}

\begin{table}[h!]
\centering
\caption{Number of lags and forecast horizons used for each dataset
in the fixed-scheme forecasting study.}
\label{lags_hotizon}
\begin{tabular}{lcc}
\toprule
\textbf{Dataset} & \textbf{Number of lags} & \textbf{Forecast horizon} \\
\midrule
Electricity Weekly & 65  & 8  \\
M4 Hourly          & 210 & 48 \\
Saugeen Day        & 9   & 30 \\
Tourism Monthly    & 15  & 24 \\
Car Parts          & 15  & 12 \\
\bottomrule
\end{tabular}
\end{table}

\begin{table}[htbp]
\centering
\caption{Comparison of forecasting accuracy across datasets and models in terms of MAE, reported as the mean with the standard deviation in parentheses. For each dataset, the lowest mean and lowest standard deviation are highlighted independently in bold.}
\label{tab:mean_mae}
\resizebox{\textwidth}{!}{%
\begin{tabular}{lrrrrr}
\toprule
\textbf{Model} & \textbf{Electricity weekly} & \textbf{M4 hourly} & \textbf{Saugeen day} & \textbf{Tourism monthly} & \textbf{Car parts} \\
\midrule
\textbf{EliMoE} & 30501.05 (1571.66) & 273.60 (7.48) & \textbf{21.18} (\textbf{0.19}) & \textbf{1809.03} (\textbf{20.63}) & \textbf{0.39} (\textbf{0.00}) \\
\midrule
\textbf{MoGU} & 31266.07 (\textbf{1435.06}) & 281.87 (\textbf{3.80}) & 22.75 (0.41) & 1897.62 (56.54) & 0.65 (0.01) \\
\midrule
\textbf{MoLE} & 30840.22 (5062.49) & 302.21 (39.69) & 23.36 (0.82) & 1816.70 (135.14) & 0.48 (0.01) \\
\midrule
\textbf{Moirai-MoE} & 59247.96 (666.71) & \textbf{240.10} (6.40) & 29.52 (0.83) & 4914.96 (28.59) & 0.42 (0.00) \\
\midrule
\textbf{PatchTST} & 35848.88 (3955.40) & 342.23 (13.78) & 23.85 (0.27) & 2236.85 (62.37) & 0.49 (0.01) \\
\midrule
\textbf{TimeMoE} & 55794.93 (--) & 301.28 (--) & 32.09 (--) & 4431.75 (--) & 0.58 (--) \\
\midrule
\textbf{DeepAR} & 54329.90 (10730.86) & 467.57 (127.61) & 28.92 (3.01) & 2032.93 (182.12) & \textbf{0.39} (\textbf{0.00}) \\
\midrule
\textbf{FFNN} & 35222.10 (3571.31) & 378.59 (45.34) & 21.39 (0.62) & 2086.13 (133.36) & \textbf{0.39} (\textbf{0.00}) \\
\midrule
\textbf{N-BEATS} & \textbf{29719.87} (4627.87) & 312.49 (45.54) & 28.12 (1.19) & 1974.64 (63.87) & 0.98 (\textbf{0.00}) \\
\midrule
\textbf{Transformer} & 66064.88 (11599.76) & 605.79 (320.80) & 28.26 (0.89) & 2743.05 (1016.36) & \textbf{0.39} (\textbf{0.00}) \\
\midrule
\textbf{WaveNet} & 63412.47 (6617.03) & 382.35 (47.49) & 22.90 (0.89) & 2302.68 (342.33) & 0.40 (0.01) \\
\bottomrule
\end{tabular}
}
\end{table}

\begin{table}[htbp]
\centering
\caption{Comparison of forecasting accuracy across datasets and models in terms of RMSE, reported as the mean with the standard deviation in parentheses. For each dataset, the lowest mean and lowest standard deviation are highlighted independently in bold.}
\label{tab:mean_rmse}
\resizebox{\textwidth}{!}{%
\begin{tabular}{lrrrrr}
\toprule
\textbf{Model} & \textbf{Electricity weekly} & \textbf{M4 hourly} & \textbf{Saugeen day} & \textbf{Tourism monthly} & \textbf{Car parts} \\
\midrule
\textbf{EliMoE} & 33309.88 (1567.32) & 330.68 (10.46) & 40.17 (0.26) & \textbf{2256.25} (\textbf{20.22}) & 0.74 (\textbf{0.00}) \\
\midrule
\textbf{MoGU} & 34443.62 (\textbf{1342.86}) & 330.84 (\textbf{4.86}) & 40.53 (0.40) & 2445.63 (85.21) & 0.84 (0.01) \\
\midrule
\textbf{MoLE} & 229619.51 (64998.78) & 1279.82 (252.64) & 44.05 (1.13) & 6352.40 (277.48) & 1.08 (\textbf{0.00}) \\
\midrule
\textbf{Moirai-MoE} & 62192.06 (578.91) & \textbf{305.07} (6.54) & 47.98 (0.98) & 6554.75 (28.69) & 0.73 (0.00) \\
\midrule
\textbf{PatchTST} & 39449.42 (3759.32) & 412.52 (14.82) & 41.59 (\textbf{0.19}) & 2826.11 (80.71) & \textbf{0.70} (\textbf{0.00}) \\
\midrule
\textbf{TimeMoE} & 58824.94 (--) & 391.25 (--) & \textbf{39.92} (--) & 5658.86 (--) & 0.77 (--) \\
\midrule
\textbf{DeepAR} & 57261.98 (10600.88) & 566.57 (137.46) & 50.02 (2.36) & 2586.55 (216.79) & 0.74 (\textbf{0.00}) \\
\midrule
\textbf{FFNN} & 38214.13 (3541.06) & 454.88 (48.60) & 41.16 (1.25) & 2586.74 (139.89) & 0.74 (\textbf{0.00}) \\
\midrule
\textbf{N-BEATS} & \textbf{32986.41} (4201.02) & 382.52 (52.59) & 49.31 (0.74) & 2551.90 (77.67) & 1.12 (\textbf{0.00}) \\
\midrule
\textbf{Transformer} & 68567.70 (11671.59) & 680.92 (321.31) & 49.19 (0.81) & 3301.36 (1098.47) & 0.74 (\textbf{0.00}) \\
\midrule
\textbf{WaveNet} & 66187.50 (6605.94) & 470.08 (54.43) & 43.82 (1.39) & 2878.62 (400.63) & 0.75 (0.01) \\
\bottomrule
\end{tabular}
}
\end{table}

\begin{table*}[h!]
\centering
\captionsetup{skip=4pt}
\caption{Effect of $\gamma$ and masking strategy (temporal masking, disjoint masking, and without masking) on EliMoE online forecasting
accuracy on the Saugeen day dataset. Values are means over 10 seeds. Bold indicates the lowest error in each row}
\label{tab:moe_gamma_masking}
%\scriptsize
\setlength{\tabcolsep}{5pt}
\renewcommand{\arraystretch}{1.1}
\begin{tabular*}{\textwidth}{@{\extracolsep{\fill}}llrrrrr@{}}
\toprule
\textbf{Setting} & \textbf{Metric} & $\boldsymbol{\gamma}=0$ & $\boldsymbol{\gamma}=0.25$ & $\boldsymbol{\gamma}=0.5$ & $\boldsymbol{\gamma}=0.75$ & $\boldsymbol{\gamma}=1$ \\
\midrule
\multirow{2}{*}{Temporal masking} & MAE & 9.40 & 8.42 & \textbf{7.61} & 8.05 & 7.98 \\
 & RMSE & 17.23 & 15.81 & \textbf{15.02} & 16.99 & 15.63 \\
\midrule
\multirow{2}{*}{Disjoint masking} & MAE & 10.09 & 8.05 & 8.12 & \textbf{7.59} & 7.98 \\
 & RMSE & 19.14 & 15.52 & 16.11 & \textbf{14.89} & 15.63 \\
\bottomrule
\multirow{2}{*}{Without masking} & MAE & \textbf{7.77} & 7.96 & 8.48 & 8.44 & 7.98 \\
 & RMSE & 15.93 & 16.08 & 17.04 & 17.01 & \textbf{15.63} \\
\midrule
\end{tabular*}
\end{table*}
\clearpage
\newpage

\begin{figure}[h!]
\centering
% You can pre-specify the width of the graph:
\includegraphics[width=0.8\textwidth]{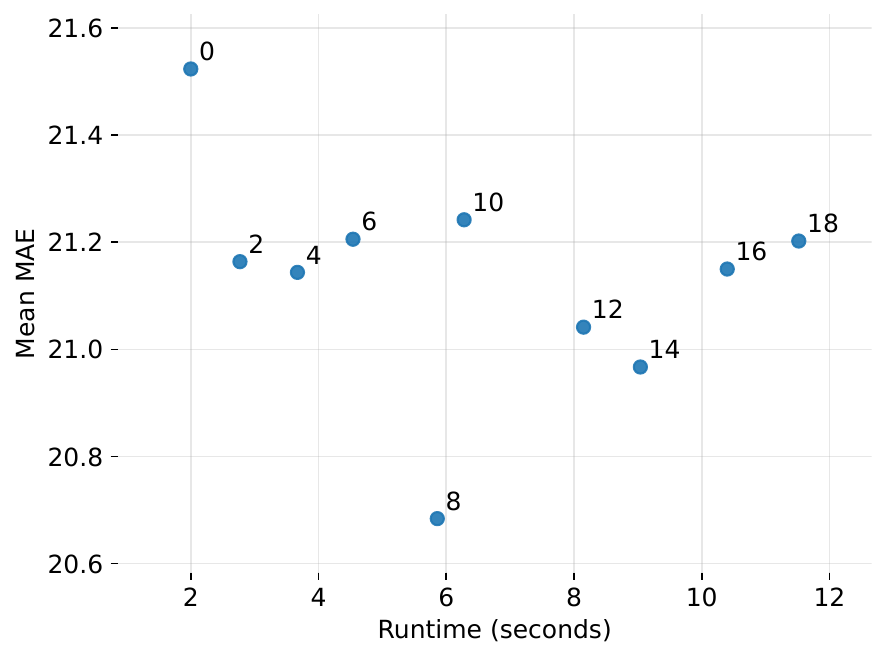}
\caption{Forecast accuracy-runtime trade-off of EliMoE for different numbers of experts. Each point represents one model configuration, with the annotation indicating the corresponding number of MLPs.}
\label{Num_expert_time}
\end{figure}

\begin{figure}[h!]
\centering
% You can pre-specify the width of the graph:
\includegraphics[width=0.7\textwidth]{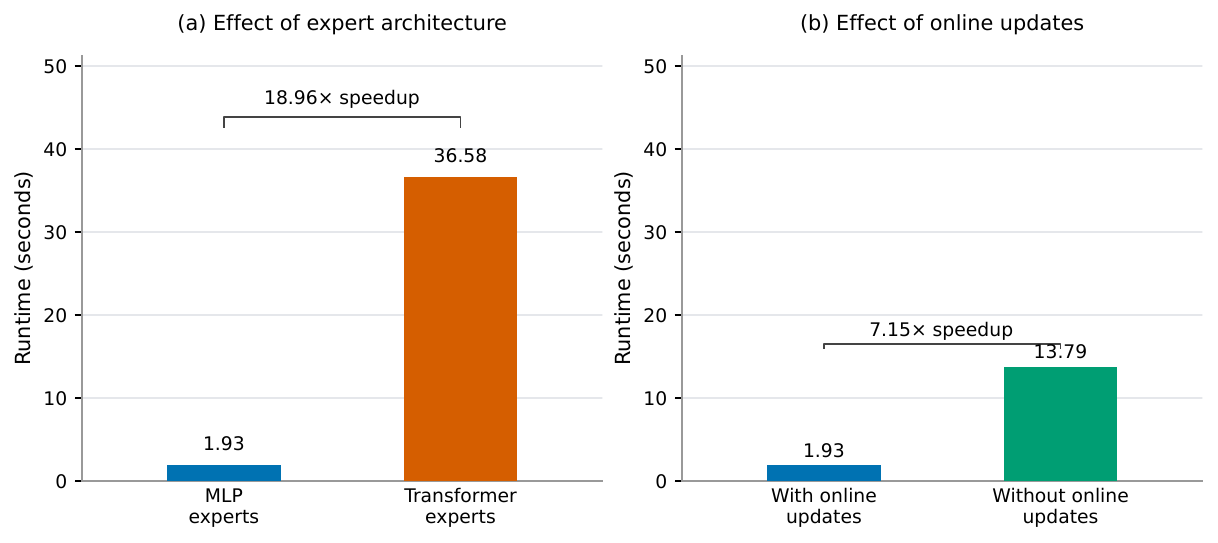}
\caption{Impact of expert architecture and partial online learning on the computational efficiency of EliMoE on the Saugeen dataset with $\gamma=0.5$.}
\label{Online_tranformer_no_online}
\end{figure}
% \begin{figure}[h!]
% \centering
% % You can pre-specify the width of the graph:
% \includegraphics[width=0.9\textwidth]{online_error_standard_deviation.pdf}
% \caption{Forecast accuracy and computational cost of FFNN, EliMoE, and MoLE on the Saugeen River Flow dataset under rolling window scheme. Bars report the mean MAE, RMSE, and runtime across ten random seeds, with error bars indicating $\pm$ one standard deviation.}
% \label{standard diviation}
% \end{figure}
\clearpage
\newpage

\section{Forecast Study}\label{forecast_study}

We conduct a fixed-window forecasting study using lagged values of the target variable as predictors. The number of lags and the forecast horizon depend on the dataset. For instance, for the Saugeen dataset, we use 9 lagged observations and a forecast horizon of 30. The lag structure and forecast horizons for all datasets follow the configurations proposed by \cite{godahewa2021monash} and are reported in Table \ref{lags_hotizon}.

EliMoE was implemented using PyTorch. We used similar hyperparameters as the FFNN model, such as the learning rate and number of epochs. However, different values were selected for other parameters such as the batch size, since the FFNN was implemented using GluonTS and required a different training configuration. Specifically, in our PyTorch implementation, each epoch the model sees the whole dataset,  whereas the GluonTS implementation of the FFNN relies on randomly sampled training  with a predefined number of batches per epoch.%Specifically, in our PyTorch implementation, each epoch corresponds to a complete pass through the training dataset, whereas the GluonTS implementation of the FFNN relies on randomly sampled training  with a predefined number of batches per epoch. %Consequently, identical batch sizes and numbers of epochs do not necessarily imply the same training budget across the two implementations.
The exact hyperparameters are summarized in the following table.

\begin{table}[h!]
\centering
\caption{Hyperparameter configuration of EliMoE under the
fixed-scheme forecasting setting.}
\label{tab:elimoe_hyperparameters}
\begin{tabular}{lc}
\toprule
\textbf{Hyperparameter} & \textbf{Value} \\
\midrule
\multicolumn{2}{l}{\textit{Model architecture}} \\
Number of linear experts & 1 \\
Number of MLP experts ($K$) & 3 \\
Hidden dimension per MLP & 60 \\
\midrule
\multicolumn{2}{l}{\textit{Training configuration}} \\
Optimizer & Adam \\
%Loss function & $L_1$ \\
Initial learning rate & $10^{-3}$ \\
Batch size & 256 \\
Maximum number of epochs & 100 \\
Weight decay & 0 \\
\midrule
\multicolumn{2}{l}{\textit{Learning rate scheduler}} \\
Scheduler & ReduceLROnPlateau \\
Reduction factor & 0.25 \\
Patience & 5 \\
Threshold & $10^{-2}$ \\
Threshold mode & Absolute \\
Minimum learning rate & $5 \times 10^{-5}$ \\
\midrule
\multicolumn{2}{l}{\textit{Early stopping}} \\
Patience & 10 \\
Minimum improvement & $10^{-2}$ \\
Monitored loss & Training objective \\
Restore best weights & Yes \\
%\midrule
% \multicolumn{2}{l}{\textit{Experimental configuration}} \\
% Number of independent runs & 10 \\
% Random seeds & 1--10 \\
% Computational device & CPU \\
\bottomrule
\end{tabular}
\end{table}

% \begin{table}[h]
% \centering
% \caption{Hyperparameter configuration for the MoE and FFNN models}
% \label{tab:hyperparameters}
% \begin{tabular}{lcc}
% \hline
% \textbf{Hyperparameter} & \textbf{MoE} & \textbf{FFNN (GluonTS)} \\
% \hline
% Learning rate & 0.001 & 0.001 \\
% Weight decay & 1e-8   & -\\
% Batch size & 256 & 32 \\
% Number of epochs & 100  & 100 \\
% Number of batches per epoch& -  & 50 \\
% Optimizer & Adam & Adam \\
% Framework & PyTorch & GluonTS \\
% \hline
% \end{tabular}
% \end{table}

%For the number of neurons, the FFNN employs 20 neurons in the first hidden layer and 20 neurons in the second hidden layer. For the MoE model, we use 40 neurons when only one MLP expert is employed, and 20 neurons for each MLP expert when three MLP experts are used.

In the rolling-window forecasting study, the Saugeen dataset is used. The last 30 observations are kept untouched as the test set, while the 30 observations preceding the test set are allocated to a validation set for hyperparameter tuning.

For all models, the hyperparameter settings from the fixed-scheme forecasting study are retained, except for a subset of hyperparameters that are tuned according to the model architecture, namely the number of hidden units, the learning rate, and the window size. Table \ref{tab:tuned_hyperparameters} summarizes the search space considered for these hyperparameters.

\begin{table}[h!]
\centering
\small
\caption{Hyperparameter search spaces for EliMoE and the benchmark
models in the rolling-window forecasting study.}
\label{tab:tuned_hyperparameters}
\begin{tabular}{lcccc}
\toprule
\textbf{Model} 
& \textbf{Neurons} 
& \textbf{Learning rate} 
& \textbf{Window size} 
& \textbf{Hidden units} \\
\midrule
FFNN 
& $[20,80]$ 
& $[10^{-5},10^{-2}]$ 
& $[730,3650]$ 
& -- \\

EliMoE (Initial window) 
& $[20,80]$ 
& $[10^{-5},10^{-2}]$ 
& $[730,3650]$ 
& -- \\

EliMoE (Updated window) 
& $[20,80]$ 
& $[10^{-4},10^{-2}]$ 
& $[1,1095]$ 
& -- \\

MoLE 
& -- 
& $[10^{-5},10^{-2}]$ 
& $[730,3650]$ 
& $[8,512]$ \\

MoGU 
& -- 
& $[10^{-5},10^{-2}]$ 
& $[730,3650]$ 
& --  \\

ADAPT-Z 
& -- 
& $[10^{-5},10^{-2}]$ 
& $[730,3650]$ 
& $[8,512]$ \\
\bottomrule
\end{tabular}
\end{table}

% \begin{table}[h!]
% \centering
% \small
% \caption{Tuned hyperparameters for benchmarks and EliMoE in the rolling-window forecasting study.}
% \label{tab:tuned_hyperparameters}
% \begin{tabular}{lcccc}
% \toprule
% \textbf{Hyperparameter} & \textbf{FFNN} & \textbf{EliMoE (Initial Window)} & \textbf{EliMoE (Updated Window)} & \textbf{MoLE}\\
% \midrule
% Number of neurons &[20, 80]   &[20, 80]   &[20, 80]& -   \\
% Learning rate &  [1e-5, 1e-2] & [1e-5, 1e-2] & [1e-4, 1e-2]  & [1e-5, 1e-2]\\
% Window size &[730, 3650]    &[730, 3650]   & [1, 1095] &[730, 3650]  \\

% Number of hidden units &-  &- & - &[8, 512]  \\
% \bottomrule
% \end{tabular}
% \end{table}
Note that all calculations for the proposed and benchmark models were conducted on a MacBook Pro (2020) equipped with an Apple M1 processor (8 CPU cores) and 16 GB of unified memory.
%%%%%%%%%%%%%%%%%%%%%%%%%%%%%%%%%%%%%%%%%%%%%%%%%%%%%%%%%%%%

\end{document}